\documentclass[conference]{IEEEtran}
\IEEEoverridecommandlockouts

\usepackage{cite}
\usepackage{amsmath,amssymb,amsfonts}
\usepackage{graphicx}
\usepackage{textcomp}
\usepackage{xcolor}
\usepackage{url}

\usepackage{tikz}
\usetikzlibrary{positioning,arrows.meta,fit,calc}

\tikzset{
  octbox/.style   = {draw, rounded corners=2pt, align=center, font=\scriptsize,
                     inner sep=3pt, minimum height=5.5mm},
  octstage/.style = {octbox, fill=black!8, minimum width=13mm},
  octspec/.style  = {octbox, fill=black!3, densely dashed},
  octpriv/.style  = {octbox, fill=black!15},
  octagent/.style = {octbox, fill=black!25, thick},
  octlabel/.style = {font=\scriptsize\itshape, align=center},
  octarr/.style   = {-{Stealth[length=1.6mm]}, semithick},
  octarrd/.style  = {octarr, densely dashed},
}

\def\BibTeX{{\rm B\kern-.05em{\sc i\kern-.025em b}\kern-.08em
    T\kern-.1667em\lower.7ex\hbox{E}\kern-.125emX}}
\begin{document}

\title{Octopus Protocol: One-Shot Hardware Discovery and Control for AI Agents via Infrastructure-as-Prompts}

\author{
\IEEEauthorblockN{Quilee Simeon}
\IEEEauthorblockA{
\textit{Massachusetts Institute of Technology}\\
qsimeon@mit.edu
}
\and
\IEEEauthorblockN{Justin Minggan Wei}
\IEEEauthorblockA{
\textit{Harvard University}\\
mwei@fas.harvard.edu
}
\and
\IEEEauthorblockN{Yile Fan}
\IEEEauthorblockA{
\textit{Harvard University}\\
yile\_fan@mde.harvard.edu
}
}

\maketitle

\begin{abstract}
Bringing a previously unintegrated device under the control of an AI agent still requires device-specific engineering: driver selection, dependency resolution, interface design, and deployment, repeated per device and per platform. We present Octopus, a hardware onboarding framework in which a coding agent, rather than a shipped integration, is the runtime that produces the required infrastructure. Given shell access and a model API key, a single bootstrap command drives the agent through a five-stage pipeline that enumerates operating-system-visible hardware, infers device identity and capabilities, generates typed Model Context Protocol tools and the hardware-facing code behind them, and activates the result as a live endpoint. A persistent daemon then maintains the result, repairing defined classes of failure in the deployment it produced. Across four hosts spanning two processor architectures, three operating-system families, and two device-access paths, identical prose specifications produced working interfaces with no per-host edits and no hand-written integration code. Five consecutive runs on the reference host completed end to end on first attempt. We report both the resulting capability and a failure mode of unattended repair loops observed over eleven hours of continuous operation.
\end{abstract}

\begin{IEEEkeywords}
agentic systems, hardware abstraction, software engineering, physical AI, robotics infrastructure, embodied AI, Model Context Protocol, self-healing systems
\end{IEEEkeywords}

\section{Introduction}
\label{sec:intro}

\begin{figure}[t]
\centering
\resizebox{\linewidth}{!}{%
\begin{tikzpicture}[
  x=1mm, y=1mm,
  hbox/.style   = {draw, rounded corners=1.2pt, align=center,
                   font=\scriptsize, inner sep=2pt},
  hspec/.style  = {hbox, fill=black!3, densely dashed},
  hagent/.style = {hbox, fill=black!22, thick},
  hcard/.style  = {hbox, fill=black!6},
  hcardk/.style = {hbox, fill=black!15, thick},
  hdaem/.style  = {hbox, fill=none, densely dashed},
  hlab/.style   = {font=\scriptsize\itshape, inner sep=1pt},
  hsym/.style   = {font=\scriptsize, inner sep=1pt},
  harr/.style   = {-{Stealth[length=1.5mm]}, semithick},
  harrb/.style  = {{Stealth[length=1.5mm]}-{Stealth[length=1.5mm]}, semithick},
  harrd/.style  = {-{Stealth[length=1.5mm]}, semithick, densely dashed},
]

\draw[densely dashed, gray] (-4,23) rectangle (92,101);
\node[hsym, anchor=north east] at (90,100) {$I$,\;\textit{infrastructure}};

\node[anchor=north west] at (-3,101) {$I = A(S,P)$};

\node[hsym, anchor=south west] at (16,91.3) {$S$,\;\textit{specification}};
\node[hspec, minimum width=56mm, minimum height=5mm] (boot) at (44,88.5)
  {\textbf{\ttfamily curl \ldots{} | bash}};
\node[hspec, minimum width=56mm, minimum height=8mm] (spec) at (44,81)
  {\textbf{\texttt{protocol/*.md}}\\[1pt]
   365 lines of prose};

\draw[harr] (spec.south) -- (44,71.5);

\node[hsym, anchor=south west] at (11,72) {$A$,\;\textit{agent}};
\node[hagent, minimum width=66mm, minimum height=13mm] (agent) at (44,65)
  {\textbf{coding-agent runtime}\\[1pt]
   harness $+$ model\\[2pt]
   \textsc{probe} $\to$ \textsc{identify} $\to$ \textsc{interface}
   $\to$ \textsc{serve} $\to$ \textsc{deploy}};

\draw[harr] (44,54) -- (44,58.5);
\draw[semithick] (11,54) -- (77,54);
\foreach \x in {11,33,55,77} { \draw[harr] (\x,54) -- (\x,50.5); }

\node[hsym, anchor=south west] at (1,54) {$P$,\;\textit{platform}};
\node[hcard,  minimum width=20mm, minimum height=12mm] (h1) at (11,44)
  {\textbf{Ubuntu}\\ x86-64, lsusb\\[1.5pt] \textit{10 tools in 4.47\,min}};
\node[hcard,  minimum width=20mm, minimum height=12mm] (h2) at (33,44)
  {\textbf{Windows}\\ WSL2, usbipd\\[1.5pt] \textit{4 tools in 3.65\,min}};
\node[hcard,  minimum width=20mm, minimum height=12mm] (h3) at (55,44)
  {\textbf{Raspberry Pi}\\ 1GB RAM\\[1.5pt] \textit{11 tools in 7.97\,min}};
\node[hcard, minimum width=20mm, minimum height=12mm] (h4) at (77,44)
  {\textbf{macOS}\\ Apple silicon\\[1.5pt] \textit{7 tools in 5.43\,min}};

\foreach \x in {11,33,55,77} { \draw[harrd] (\x,32) -- (\x,37.5); }
\node[hdaem, minimum width=84mm, minimum height=7mm] (daem) at (44,28.5)
  {persistent daemon: \textsc{watch} $\to$ \textsc{heal}};

\draw[harrd] (daem.west) -- (-1,28.5) |- (agent.west);

\end{tikzpicture}%
}
\caption{Octopus overview, the instantiated execution model. Identical specifications, executed on four machines, produced
working hardware interfaces via MCP endpoints with no hand-written integration code.}
\label{fig:hero}
\end{figure}

Personal computers once required users to install vendor-specific drivers before newly connected hardware became usable; modern operating systems absorbed that burden behind standardized abstractions and plug-and-play. Agentic AI is at the analogous earlier stage. Today's coding agents can help users control physical devices, yet bringing a previously unintegrated device online still requires substantial device-specific engineering: discovery, dependency selection, interface generation, debugging, and deployment, guided step by step by the user. This paper asks whether that onboarding can itself become plug-and-play: whether a coding agent executing a general protocol against the target machine can generate and maintain the required infrastructure automatically.

The cost of the status quo is documented, not hypothetical. Producing the control stack that agentic-robotics systems presume is expensive: MoveIt's own 105-user study reports a mean of 1.5 hours from robot description to motion planning, with only 28\% of users rating setup positively~\cite{coleman2014moveit}. We call this the glue-code tax, and it is the problem Octopus targets.

Existing interface-generation work starts one step later. Across the systems we survey in \S\ref{sec:related}, agent-facing hardware interfaces are either \emph{hand-authored}, \emph{compiled from a supplied machine-readable contract}, or \emph{generated above an already-integrated platform}, and all three assume the device-specific artifact that onboarding must produce. Beginning from nothing but what the operating system exposes is the missing step, and the step this paper studies.

\paragraph{Octopus} In Octopus\footnote{Project site and implementation: https://octopusprotocol.ai/ and https://qsimeon.github.io/octopus-hw/} the coding agent \emph{is} the onboarding mechanism: nothing device-specific is shipped, and the interface is built on the machine that will use it. One bootstrap command, given ordinary shell access and a model API key, carries the agent through the five-stage pipeline of \S\ref{sec:design} to a live Model Context Protocol (MCP) endpoint any client can call, after which a persistent daemon keeps that endpoint working. We formalize the execution model as
\begin{equation}
I = A(S,P),
\label{eq:iasp}
\end{equation}
in which the infrastructure $I$ is what agent $A$ produces by executing portable prompt-level specifications $S$ on target platform $P$. The portable artifact is $S$; everything platform-specific in $I$ is materialized at deployment time. Fig.~\ref{fig:hero} shows (\ref{eq:iasp}) instantiated on the four hosts evaluated here. Rather than shipping a fixed integration as the software, Octopus treats prompt directives as a portable onboarding protocol executed by an agent that is no longer a programming assistant but the runtime of the infrastructure itself. \textbf{In this model, a living coding agent is the software; the code is just the artifact.}

Onboarding here is \emph{one-shot} in the interaction sense: bounded user involvement, a single bootstrap action, rather than a single model invocation. The agent may retry and self-correct and keeps working after deployment, but the user never orchestrates the integration step by step.

\paragraph{Contributions} In the terms of (\ref{eq:iasp}) this paper contributes four things: \textbf{[S, P]} an onboarding architecture that begins from device-agnostic prompt directives and OS-visible hardware rather than a supplied contract, SDK, or wrapper; \textbf{[A]} a deployment-time runtime model in which the agent acts as compiler, deployer, and maintainer of platform-specific infrastructure; \textbf{[I]} the automatic generation of live, typed MCP infrastructure immediately usable by downstream agents; and \textbf{[I]} a persistent daemon that turns the same runtime on its own output, repairing enumerated classes of dependency, connectivity, and generated-code failure after deployment.

Our novelty claim is compositional: individual ingredients have strong neighbors~\cite{mastouri2026autompc, guan2026roboneuron, guo2025sensormcp} and no single stage is unprecedented, but no prior system composes the whole sequence. \S\ref{sec:related} sets each ingredient beside its nearest neighbor; that composition is what this paper evaluates.

\section{Related Work}
\label{sec:related}

Language-model robotics established that agents can plan and program robots~\cite{liang2023cap,ahn2022saycan}, and vision-language-action models scaled the premise~\cite{zitkovich2023rt2,kim2024openvla}. All presuppose that a control stack for the target embodiment already exists, and the newest agentic frameworks document that assumption rather than hide it: RoboClaw's robot developer specifies system configuration, MCP tools, and skills by hand~\cite{li2026roboclaw}, and CaP-X supplies perception and control primitives with schemas at every abstraction tier~\cite{fu2026capx}. Operator-layer systems inspect and govern \emph{existing} ROS graphs rather than producing them~\cite{royce2024rosa,cardenas2026rosclaw}; their capability discovery is introspection of what running nodes advertise, not generation.

Work that builds the interface layer itself, mostly around the Model Context Protocol~\cite{anthropic2024mcp}, divides by where the interface comes from. MCP standardizes how a client discovers and invokes the tools a running server advertises, but says nothing about how a server for a physical device comes to exist: a standard target, not a standard producer. \emph{Hand-authored}: IoT-MCP's per-sensor edge servers~\cite{yang2025iotmcp}, DCP's capability manifest and validating host-side bridge~\cite{yang2026dcp}, and RCP's handlers wrapped in advance~\cite{lee2025rcp}: in each a human writes the device contract and the system governs how it is used. \emph{Compiled from a contract that already exists}: AutoMCP from OpenAPI specifications~\cite{mastouri2026autompc}, RoboNeuron from ROS message schemas~\cite{guan2026roboneuron}, TaskSense from source code and manufacturer API descriptions~\cite{liu2025tasksense}. \emph{Generated above an integrated platform}: SensorMCP builds task-specific sensor tools for smart cameras, though onboarding the camera is outside its scope~\cite{guo2025sensormcp}. Octopus differs in its input rather than its output: it begins from OS-visible hardware for which no machine-readable contract is supplied, so the capability description must itself be inferred before any interface can be generated. ROS~2 bounds the contrast from below, supplying typed transport and discovery of graph participants while its drivers and interface definitions remain engineering inputs~\cite{macenski2022ros2}.

The generative mechanism has a lineage. Formal synthesis reached device drivers with Termite-2, at roughly a week per specification and presuming the device is fully modeled in advance~\cite{ryzhyk2014termite}, a correctness-for-generality trade Octopus takes in the opposite direction. LLM software engineering supplies the generator and the harness--model pattern~\cite{yang2024sweagent,li2026skilled}, and Infrastructure-as-Code frames the deployment half~\cite{jana2026terraformer,stoica2024specs}. Autonomic computing supplies the maintenance loop~\cite{kephart2003autonomic}, which agentic repair has since made generative~\cite{bouzenia2025repairagent}. AIOS and MemGPT place an operating layer \emph{beneath} agents, holding tools and interfaces fixed~\cite{mei2025aios,packer2023memgpt}; Octopus inverts that direction, putting the agent where the driver installer and the build system used to be.

Every neighbor above therefore begins from a human-supplied device-specific artifact; Octopus is, to our knowledge, the only system whose device-specific input is the OS-visible device itself. Several neighbors provide what it lacks, notably DCP's per-call validation, and are complements, not competitors.

\section{System Design}
\label{sec:design}

\begin{figure}[b]
\centering
\resizebox{\linewidth}{!}{%
\begin{tikzpicture}[node distance=3mm and 6mm]
  \node[octspec, minimum width=20mm] (spec) {spec $s_i$\\ (\texttt{protocol/*.md})};
  \node[octbox, below=of spec, minimum width=20mm] (prev) {previous artifact\\ $o_{i-1}$ (file path)};
  \node[octbox, right=of spec, minimum width=18mm, yshift=-5mm] (prompt) {prompt\\ builder};
  \node[octagent, right=of prompt, minimum width=20mm, minimum height=12mm] (agent)
    {agent $A$\\ \scriptsize harness $\times$ model\\ \scriptsize (swap points)};
  \node[octstage, right=of agent, minimum width=18mm] (out) {declared output\\ file $o_i$};

  \node[octbox, below=2.5mm of agent, minimum width=22mm, densely dashed] (retry)
    {retry: spec by reference\\ (self-edit) + log tail};
  \node[octbox, minimum width=18mm] (check) at (out |- retry) {exists \&\\ non-empty?};
  \draw[octarr] (spec) -| (prompt);
  \draw[octarr] (prev) -| (prompt);
  \draw[octarr] (prompt) -- (agent);
  \draw[octarr] (agent) -- (out) node[midway, above, octlabel] {writes};
  \draw[octarr] (out) -- (check);

  \draw[octarr] (check.west) -- (retry.east)
      node[midway, above=3.6mm, octlabel] {no};
  \draw[octarrd] (retry.north) -- (agent.south);
  \draw[octarr] (check.east) -- ++(6mm,0)
      node[right, octlabel, align=left] {yes:\\ $o_i \to o_{(i+1)-1}$\\ next stage};
  \node[draw, rounded corners=3pt, dotted, fit=(prompt)(agent)(out)(check)(retry),
        inner sep=3.5mm, label={[octlabel]above:{orchestrator (budgeted, fail-stop; no semantic validation)}}] {};
\end{tikzpicture}%
}
\caption{One iteration of the orchestration loop.}
\label{fig:architecture}
\end{figure}

\subsection{Overview and Design Goals}
\label{sec:design:goals}
\label{sec:design:assumptions}

The system ships no device-specific integration code as executable software. It ships \emph{specifications} (markdown documents describing what each onboarding stage must accomplish) and a thin orchestrator that sequences a coding agent through them on the target host. The agent reads each specification, inspects the machine it is running on, writes whatever platform-specific code the specification calls for, and advances the pipeline by producing a declared output file. The generated infrastructure is a \emph{product} of deployment, not an input to it.

The design goals are the contributions of \S\ref{sec:intro} read as requirements, plus one boundary that shapes everything below: \textbf{operating-system visibility does not guarantee semantic identifiability}. Octopus degrades explicitly, not silently, when a host exposes a device's transport but not its meaning. Autonomy is bounded throughout by per-attempt and per-stage budgets under a fail-stop rule.

We assume shell access under an ordinary user account, network access to a language model and package registries, an operating system that enumerates hardware conventionally, and physical setup by the user. We do \emph{not} assume a device SDK, ROS interface, capability manifest, or control schema. The agent may discover and reuse public libraries at deployment time (choosing the right one is part of the problem) but no device-specific integration code ships as executable software. Per-call gating, formal correctness guarantees, and multi-tenant access control are non-goals (\S\ref{sec:design:safety}).

\subsection{Semantic Observability and the Two-Layer Model}
\label{sec:design:observability}

A device may be fully visible to the OS while exposing too little to determine what object it is. A host may report only

\texttt{/dev/ttyACM0}\quad\texttt{USB serial device}\quad\texttt{VID:PID}

which does not distinguish a robotic arm from a sensor board, a 3D-printer controller, or a custom microcontroller application. No amount of further reasoning recovers semantics absent from the evidence. We call the degree to which OS-visible evidence, descriptors, drivers, identifiers, and public documentation reveal a device's identity and usable capabilities its \textbf{semantic observability}, and we design across that spectrum rather than pretend it does not exist.

The boundary induces two layers of specification. \textbf{Layer 1} is five device-agnostic specifications (\textsc{probe}, \textsc{identify}, \textsc{interface}, \textsc{serve}, \textsc{deploy}; 365 lines of markdown) describing how to onboard hardware \emph{as a class of problem}, with no knowledge of any product. For semantically observable hardware this layer operates alone, and that is the strongest claim the system supports. \textbf{Layer 2} is curated device profiles for our reference hardware (an SO-ARM101 arm and a custom camera rail) held in a separate directory so the two layers are distinguishable. They supply what the operating system cannot: identity, constraints, calibration conventions, command grammars. We present them as exactly that, a disclosed semantic prior, not evidence of contract-free discovery.

Two mechanisms connect the layers and both are conditional. Identification recognizes a profiled device only when the host's own evidence corroborates it, so a profile cannot manufacture hardware that is not there; and \emph{privileged framework steps}, switched on in configuration, run once after the pipeline to verify and if necessary rewrite a profiled device's tools. The core evaluation runs with the steps disabled and no profiled hardware attached, so neither is active.

What matters is \emph{what kind of thing a profile is}: prompt-level text executed by the same runtime, not a driver, a pre-built server, or integration code. It answers ``what is this thing?'' in natural language, which is why its location is incidental: a profile committed to the repository and the same words typed at discovery time are the same input to the same mechanism, and eliciting it through a question is the extension we chart in \S\ref{sec:disc:future}. Even for profiled devices the implementation, tool definitions, dependency selection, deployment artifacts, and later repairs are all agent-generated. Octopus does not always eliminate device-specific knowledge; it moves device-specific \emph{implementation} out of fixed code and into executable text.

\subsection{The Three Hardware Tiers}
\label{sec:design:tiers}

Semantic observability is a spectrum, and at every point the practical question is the same: \emph{how much must a person say before the agent can proceed?} Table~\ref{tab:tiers} answers at three points, and the answer is always measured in sentences: in no tier does anyone write, port, or wrap integration code. The step from Tier~2 to Tier~3 is the instructive one: a name suffices because everything downstream of it is public and the agent can retrieve it, and a name buys nothing for a device that exists in a single instance, so the sentence has to become a description of what the thing is and how it is commanded.

\begin{table}[htbp]
\caption{The three hardware tiers: how much a person must say.}
\begin{center}
\footnotesize
\setlength{\tabcolsep}{3pt}
\begin{tabular}{|l|l|l|l|}
\hline
\textbf{Tier} & \textbf{\textit{Host reveals}} & \textbf{\textit{Person says}} & \textbf{\textit{Rest from}} \\
\hline
\textbf{1} camera, audio & identity + capability & nothing & the platform \\
\hline
\textbf{2} SO-ARM101 & transport only & a \emph{name} & public docs \\
\hline
\textbf{3} camera rail & transport only & a \emph{description} & nothing else \\
\hline
\end{tabular}
\label{tab:tiers}
\end{center}
\end{table}

\subsection{Architecture}
\label{sec:design:architecture}
\label{sec:design:specformat}

Octopus comprises four components with strict separation of concerns (Fig.~\ref{fig:architecture}). \textbf{The orchestrator} (724 lines of Python) is the only conventionally executed program, and it is deliberately unintelligent: it renders prompts, spawns the agent, sequences stages, and checks files. \textbf{The harness} is an off-the-shelf coding-agent CLI supplying a ReAct-style tool loop over shell, files, and web search; \textbf{the model} is any LLM it can reach. Each is one configuration line, making harness, model, and \textbf{specifications} three independently swappable layers: the concrete instantiation of (\ref{eq:iasp}), where replacing $A$ requires no change to $S$. The whole human-written core is 2{,}921 lines and contains no device database, driver registry, hardware-access implementation, or platform-specific execution path; everything a deployed host runs is materialized by the agent at deployment time into a separate output tree, alongside a unified execution log the daemon later reads.

\paragraph{Output contracts are the type system} Because the orchestrator deliberately does not validate content, the schemas stated in specifications are the only interface definitions in the system: \textsc{identify}'s capability names are the contract that \textsc{interface} must reproduce exactly, and \textsc{watch} and \textsc{heal} outputs carry fixed schemas because the daemon branches on their fields. Constraints in the same documents encode platform knowledge as text rather than code: per-bus concurrency locks, a capture-and-release camera discipline, and a logging wrapper on every generated tool, all engineering judgment that in a conventional stack would be a code-review checklist.

\paragraph{Markov context propagation} Each stage $i$ receives exactly two inputs: its own specification $s_i$ and the artifact of the preceding stage $o_{i-1}$, so $o_i = A(s_i, o_{i-1}, P)$ and the composition over stages yields $I$. No conversation history crosses stage boundaries, so context is bounded by construction however long the pipeline, every stage is independently re-runnable from its predecessor's artifact (the basis of resume and re-discovery), and the file boundary is the auditable record of what each stage knew. The daemon applies the same rule to perception: two frames plus a rolling summary, compressed rather than accumulated. Generation and repair use the full agent; housekeeping judgments such as compressing an overgrown log use a bare completion call with a token cap.

\paragraph{Self-editing specifications} A first attempt embeds the stage specification in the prompt. A retry instead \emph{points at the specification file}, permits the agent to edit it if the failure traces to a defect there, and injects a bounded tail of the failed attempt's log. Edits are detected by version-control diff and logged. The specification is therefore not merely the program but a mutable part of the system's state: the mechanism by which a failing deployment repairs its own instructions within the run.

\subsection{Lifecycle and Bounded Self-Healing}
\label{sec:design:lifecycle}
\label{sec:design:failure}

A single bootstrap command installs the harness and its runtime, clones the repository, provisions credentials, and launches the orchestrator. One-shot onboarding begins there: after that action the user does not steer discovery, library selection, code generation, debugging, or deployment, and no pipeline stage stops to ask anything.

The five stages then run. \textsc{Probe} enumerates everything the operating system will reveal, using only tools already present and labeling network neighbors as \emph{context} rather than controllable devices. \textsc{Identify} infers what each endpoint \emph{is} from descriptors, drivers, identifiers, and web search over the \emph{combination} of clues, annotating identity, capabilities, a confidence score, and the method used, so the inference is auditable; its guiding rule is to identify the product the chip belongs to rather than the chip, and generic passthrough capabilities are prohibited as tool material. \textsc{Interface} turns capabilities into JSON-Schema-typed tool definitions under a bounded budget. \textsc{Serve} writes one self-contained FastMCP server implementing every tool against real hardware, and \textsc{deploy} installs dependencies, syntax-checks, selects a port, writes the connection descriptor, client snippets, and a launcher, then starts the server detached.

After deployment the daemon takes over, and the pipeline does not run again except through re-discovery. \textsc{Watch} hands the agent the unified log (every generated tool call is logged by construction, so the daemon is not blind) and, when perception is enabled, the current frame and rolling summary, and requires a health report in a fixed schema. \textsc{Heal} receives that report and the server and acts under an enumerated playbook: missing dependency $\to$ install; broken tool $\to$ rewrite the function; corrupted file $\to$ repair syntax; dead process $\to$ restart through the generated launcher; and \emph{device moved}, where the device is present but re-enumerated elsewhere $\to$ do \emph{not} rewrite correct code, but report the classification so the daemon triggers re-discovery, which re-runs \textsc{probe} and \textsc{identify} and patches the live server without regenerating the \textsc{serve} and \textsc{deploy} artifacts. \textsc{Heal} never deletes tools, backs up before modifying, and syntax-checks afterwards; liveness restarts need no agent call. What the design does \emph{not} claim is equally structural: failures outside the enumerated classes have no playbook entry, and the report schema's \texttt{failed} status and \texttt{fatal} severity exist to surface them for human attention rather than absorb them.

Two channels of evidence: The log records what the \emph{software} did; the frame and its rolling summary record what the \emph{world} looks like. Self-managing systems conventionally monitor only the first, because the second has no schema (there is no exception type for ``the gripper is open when the state says closed''), and a general-purpose model in the monitor makes it usable without one, since the frame is simply more evidence in the same prompt. This is what makes the loop \emph{proprioceptive} rather than merely instrumented, and how a deployment can notice that its software is nominal and its hardware is not. The channel is optional: without it, \textsc{watch} proceeds on logs alone and records that physical state was unavailable rather than inferring it.

\subsection{Safety and Threat Boundary}
\label{sec:design:safety}

Deployment scope is chosen rather than fixed. \emph{Local} uses a standard-I/O transport that opens no socket, only agents on the same machine can call the server; \emph{intranet}, the launcher's default, serves HTTP on all interfaces, which is what makes a rig usable from a second machine; \emph{internet} adds an outbound tunnel with a rotating address and one-command revocation.

Within any scope there is no per-call authorization (reachability is authorization), and where tools drive motors that includes causing physical motion. Profiles encode device-protective constraints such as clamped ranges and thermal limits, but those bound damage, not authority. This is a scoping decision rather than an oversight: because what we emit is an ordinary MCP server, a per-call gating layer composes in front of it without changing anything described here.

\section{Evaluation}
\label{sec:eval}
\label{sec:eval:setup}

The evaluation is asymmetric by design. Tier~1, semantically observable hardware, both Layer-2 mechanisms disabled, carries the quantitative claims; Tiers~2 and~3 are profile-assisted case studies, each reported with its supplied prior itemized.

All experiments run one frozen version of Octopus: orchestrator, all nine specifications, installer, and configuration. The model is \texttt{google/gemini-3-flash-preview} through OpenRouter for every quantitative run. Four hosts: the \emph{reference host} is an Ubuntu workstation (i7-6700T, 32\,GB); portability is established on an M1 macOS laptop, a 1\,GB Raspberry Pi~3B, and Windows~11/WSL2. A Microsoft LifeCam Studio is attached to every host, each host's onboardable inventory is fixed, and the Tier-1 condition removes both routes by which device-specific knowledge could reach the pipeline: privileged steps disabled in configuration, device-expectation fields cleared, no profiled hardware attached. Every Tier-1 tool is produced by the five general specifications alone.

\begin{table}[htbp]
\caption{Tier-1 onboarding, 5 consecutive runs on the reference host.}
\begin{center}
\scriptsize
\setlength{\tabcolsep}{3pt}
\begin{tabular}{|l|r|r|r|r|r|r|}
\hline
\textbf{Metric} & \textbf{\textit{R1}} & \textbf{\textit{R2}} & \textbf{\textit{R3}} & \textbf{\textit{R4}} & \textbf{\textit{R5}} & \textbf{\textit{Med.}} \\
\hline
Completed end-to-end & yes & yes & yes & yes & yes & 5/5 \\
\hline
Pipeline retries & 0 & 0 & 0 & 0 & 0 & 0 \\
\hline
Total time (s) & 204 & 277 & 308 & 268 & 233 & 268 \\
\hline
Generated tools & 10 & 6 & 5 & 12 & 13 & 10 \\
\hline
Devices enumerated & 154 & 103 & 191 & 122 & 64 & 122 \\
\hline
\quad annotated & 9 & 12 & 8 & 8 & 8 & 8 \\
\hline
\end{tabular}
\label{tab:rq1}
\end{center}
\end{table}

\subsection{Contract-Free Onboarding Produces Working Interfaces}
\label{sec:eval:rq1}
\label{sec:eval:rq3}

Five fresh runs on the reference host completed all five stages on first attempt, with zero retries across twenty-five stage executions and zero human actions after invocation, deploying live HTTP endpoints in 204--308\,s, median 268 (Table~\ref{tab:rq1}). A run succeeds when the pipeline completes and the deployed endpoint answers \texttt{tools/list}.

No device profile contributed. The camera was identified from its USB descriptor alone, the archived output recording \texttt{identifica\-tion\_method: descriptor\_driver} at confidence 1.0 with no framework hint: semantically observable hardware onboarded with no prior supplied by anyone.

\paragraph{Coverage} Against a pre-registered inventory of \textbf{eight} onboardable devices, run~1 delivered at least one working tool for four: \textbf{50\,\%}, or 67\,\% once the Bluetooth and Wi-Fi interfaces, which \textsc{identify} deliberately classified \texttt{infrastructure}, are scored out of scope. Storage is a genuine miss, and instructive: the same specifications annotated block storage on the Raspberry Pi and not here. \textsc{probe}'s reach is far wider than the inventory it is scored against: 154 entries on the reference host, UART nodes, loop devices, and in-range Wi-Fi networks, which the pipeline narrowed to 9 annotated and 10 tools emitted, a filtering no human wrote.

\paragraph{The interfaces are correct, not merely present} All 46 tools generated across the five runs declare an \texttt{input\_schema}, and all 46 declare it as a JSON Schema object type; all five servers then registered their tool sets and served them over HTTP, which a malformed schema would have prevented at framework registration. No tool addressed a device \textsc{probe} had not observed, and none exposed a generic passthrough (raw byte, serial, or exec), which the specifications prohibit. Two tools in run~1 were duplicate paths to the same capture, redundant rather than incorrect. Camera tools were invoked over the deployed endpoint as an external client rather than by importing the module, writing frames at 1920$\times$1080; the stronger evidence is that the \textsc{watch} stage subsequently \emph{described} those frames, since a capture tool returning a schema-valid image of uniform black would satisfy every structural check while being functionally dead.

\subsection{One Specification, Many Implementations}
\label{sec:eval:rq2}
\label{sec:eval:rq5}

The same frozen specifications ran unmodified on all four hosts with the common camera attached, and every session completed: five of five on the reference host, one of one elsewhere, with a single pipeline retry on the Pi and on macOS. The hosts span two processor architectures, three operating-system families (Linux, Windows, Darwin), and two device-access paths, one of which delivers USB devices to the kernel over IP rather than over a physical bus.

What differed was everything below the specification. The agent resolved a different dependency set on each host, and tool-set composition followed the platform rather than the specification: GPIO tools on the Raspberry Pi, described nowhere in any specification and impossible on an x86 host, and display and audio tools only on the MacBook. Host-specific tools are portability's signature rather than its failure.

\paragraph{macOS is where a hardcoded implementation would have failed} Three of the four hosts run Linux, so a Linux-specific implementation would have passed on all of them. Darwin is the discriminating case, and the generated artifacts are Darwin-native throughout: \textsc{probe} enumerated through \texttt{system\_profiler} and \texttt{ioreg} rather than \texttt{lsusb}, audio was implemented against CoreAudio rather than ALSA, and serial devices appear under \texttt{/dev/cu.*} rather than \texttt{/dev/tty*}. None of those names appears anywhere in any specification. The specifications describe what enumeration and capture must \emph{accomplish}; the platform-appropriate mechanism was selected at deployment time from the evidence the host actually presented.

\paragraph{The same variance appears across runs, and it is ordered} Five runs on one host, identical in every controlled respect, separate into layers that behave differently. Completion is stable: five of five on first attempt. Identity is stable: the camera resolved to a Microsoft LifeCam Studio in all five at confidence 1.0 by descriptor match, the audio controller to the Intel HDA codec in all five, with no misidentifications and nothing hallucinated across twenty-five stage executions. A three-capability core (camera still capture, host audio playback, host audio capture) appears in every run. Above that core, composition varies: tool counts of 10, 6, 5, 12 and 13, different implementation libraries chosen for identical work, and, most freely of all, 41 distinct tool names across five runs. MCP specifies \texttt{tools/list} as the means by which a client learns what a server offers, so a conformant client absorbs that last layer completely. The ordering is the result: what the system must get right it got right every time, and what varies grows steadily less consequential with distance from the hardware.

\subsection{The Semantic Boundary}
\label{sec:eval:tier2}
\label{sec:eval:tier3}

\paragraph{Tier 2: a name, and what a name does not cover} With the SO-ARM101 on the Pi and the arm profile active, the pipeline deployed arm tools alongside camera tools, and the arm step verified joint motion physically rather than accepting a wire-level acknowledgment. Identification was evidence-gated rather than profile-asserted: the rule fires only when \textsc{probe} surfaces a CH340 (\texttt{1a86:7523}) or CH343 (\texttt{1a86:55d3}) USB-serial chip \emph{and} the surrounding data names Feetech, STS3215, LeRobot, or SO-ARM100/101; absent that corroboration the specification directs the stage to record \texttt{category: unknown} rather than guess.

The profile we shipped also carries facts true of this rig rather than of the product: servo~1 hardware-disabled, motor IDs re-flashed after a swap, a thermal rest policy for joint~2, none of which follows from the name. Tier~2 as evaluated therefore sits between the name-only ideal and Tier~3's full description, closer to the former.

\paragraph{Tier 3: a description, where no documentation exists} Our bespoke device is a two-axis camera rail: steppers on a lead-screw gantry driven through an A4988 stage by a microcontroller running firmware we wrote. Its identifier resolves to a generic board, and no SDK, datasheet, or repository describes its command grammar, because the device exists in a single instance. It also makes the control claim unambiguous: left alone the firmware runs a pre-written sweep, and when a USB host attaches the sweep yields and the firmware listens, so autonomous and commanded motion are visibly distinct. The profile has the arm's shape but longer: what the device is, how it is addressed, which commands it accepts, and the limits within which commanding it is safe. It remains prose, with no schema, SDK, or wrapper involved. On the reference host with camera, arm, and rail attached simultaneously, the pipeline completed all five stages on first attempt in 286\,s, within the range of the Tier-1 runs on the same machine despite three times the hardware, and deployed seven tools. The rail was identified at confidence 1.0 against its profile, three typed tools mirrored its command grammar exactly (\texttt{rail\_move\_x}, \texttt{rail\_move\_y}, \texttt{rail\_home}), and the rail was commanded through the deployed interface and moved.

\begin{table}[htbp]
\caption{Repair episodes across all reported deployments.}
\begin{center}
\footnotesize
\setlength{\tabcolsep}{3.5pt}
\begin{tabular}{|p{0.34\linewidth}|r|p{0.47\linewidth}|}
\hline
\textbf{Playbook class} & \textbf{\textit{Ep.}} & \textbf{\textit{Outcome}} \\
\hline
Missing / broken dependency & 1 & Diagnosed and repaired in 58\,s; verified by capture \\
\hline
Dead server, silent crash loop & 5 & Converged at episode 5; healthy 2\,h\,17\,m after \\
\hline
Device moved $\to$ re-discovery & 3 & Re-onboarded unattended, 355\,s in run~4 \\
\hline
Broken tool implementation $\to$ rewrite & 5 & Replacements strictly more robust than originals \\
\hline
\emph{Outside the playbook:} policy authoring under telemetry & 19 & First action correct; monotone escalation thereafter \\
\hline
\textbf{Total} & \textbf{33} & \\
\hline
\end{tabular}
\label{tab:rq4}
\end{center}
\end{table}

\subsection{Persistent Operation and Its Failure Mode}
\label{sec:eval:rq4}
\label{sec:eval:longitudinal}

We report repair observationally rather than by injection: there was no fault-injection campaign, so there is no detection rate and no mean time to recovery. Table~\ref{tab:rq4} is instead the complete population of episodes these deployments produced, every one arising unprompted, classified against the playbook of \S\ref{sec:design:failure}. An injection study would give rates but would measure the daemon only against faults we thought to create, and the failure that mattered most here is one we would not have thought to inject.

\paragraph{Diagnosis} The reference host supplied a dependency failure unprompted: \textsc{deploy} installed \texttt{opencv-python}, but every camera tool then raised \texttt{module 'cv2' has no attribute 'VideoCapture'}. The first watch cycle caught it 13\,s into the loop, named the three affected tools, and proposed ``headless vs full version conflicts,'' the correct diagnosis for this symptom. The repair force-reinstalled the headless build and restarted; the next invocation succeeded, \textbf{58\,s} from detection, no human involvement. No injection study could have produced this episode: the package installed successfully and the schema was valid, so every check the harness performs passed. What caught it was an agent reading a log.

\paragraph{Re-discovery} Two runs exercised the branch that couples healing back into discovery. In run~4 \textsc{interface} \emph{omitted the camera entirely}; the daemon classified the fault \emph{device moved}, re-ran \textsc{probe} and \textsc{identify}, and patched the regenerated tool into the live server 355\,s after detection. Neither that fault nor run~5's was literally a device moving, yet re-discovery is the right response to a stale model of the hardware, a loose classification reaching the right action.

\paragraph{Eleven hours unattended} After the Tier-2 pipeline completed the daemon ran unattended on the arm rig for 11\,h\,17\,m at the production 300\,s interval: 24 episodes in three regimes. It was quiescent for 4\,h\,05\,m; five episodes of convergent repair then addressed a fault worth naming, in which the generated server logged to a path the watch specification did not read, so the daemon observed not an error but an \emph{absence} and correctly inferred a silent crash loop, a materially harder inference than reacting to a raised exception, and one no specification enumerates.

The third regime is the paper's sharpest negative result. Over 2\,h\,36\,m and 19 episodes the agent met a genuine thermal signal from the elbow servo by authoring a torque ceiling, retuned PID coefficients, and an enforced duty cycle, a thermal policy no SDK ships, written at runtime for hardware it had never seen. It then tightened that policy monotonically and never revised downward, each episode inheriting the last through the rolling summary, until the diagnosis had widened from one hot joint to sheared gears, structural micro-fractures, and a \texttt{FIRE\_RISK} status that zeroed three joints and latched a terminal lock. It reported healthy for the final 1\,h\,27\,m, because every joint able to report a fault had been disabled. Almost none of the escalation was supported by an available signal: the bus reports position, load, voltage, and temperature, and nothing else. The camera watched the arm throughout without corroborating any of it. \S\ref{sec:disc:limitations} takes up why, and what follows from it.

\section{Discussion and Limitations}
\label{sec:limitations}
\label{sec:disc:results}

\label{sec:disc:limitations}

Three observations from the record seem durable. \emph{Semantic observability, not code generation, is the binding constraint}: generation never failed in the five reference runs, and the failures that mattered were identity failures; across the tiers of \S\ref{sec:eval:tier2} the code does not get harder, only the sentence a person must supply gets longer. \emph{Variance is stratified, and the ordering is favorable}: what must be right was right every time, and the layers that move sit progressively further from the hardware (\S\ref{sec:eval:rq5}), so a generative middle layer is compatible with a stable contract defined over identity and capability rather than strings. \emph{A living runtime writes policy that no library can ship}: the thermal mitigation of \S\ref{sec:eval:longitudinal} addressed no missing primitive (the SDK exposes those registers) but a missing policy, which depends on facts that do not exist when a library is written. Integration labour does not vanish under $I = A(S,P)$; it moves up a level.

The escalation of \S\ref{sec:eval:longitudinal} is the sharpest limitation, and it is architectural: the daemon's memory is prose the agent both writes and reads, with no operation that retires a conclusion, so a monotone ratchet is that arrangement's equilibrium rather than a surprise, benign here only by direction. The mitigation is cheap, because a ground-truth channel already existed and disagreed throughout but entered the prompt as a peer of the accumulated prose rather than as an authority over it; privileging direct observation, requiring each claim to cite its reading, expiring claims unless re-evidenced, and gating irreversible declarations behind human confirmation are all specification-level changes, none implemented here. The remaining bounds are stated rather than solved: reachability is authorization (\S\ref{sec:design:safety}); generated code is validated but never verified~\cite{ryzhyk2014termite,stoica2024specs}; healing covers the enumerated classes only; profiles are human-authored; and four hosts, small $N$, and a hosted model leave replication exposed to provider drift.

\section{Conclusion}
\label{sec:conclusion}

Bringing a physical device under agent control still begins, almost everywhere, with a human writing the integration. This paper asked whether that step can be delegated to the same kind of agent that will ultimately use the device. Octopus answers by construction: a five-stage, prompt-specified pipeline in which a coding agent discovers OS-visible hardware, infers what it can do, generates typed MCP tools and the code behind them, deploys them as a live endpoint, and then, as a persistent daemon invoking the same runtime, repairs defined classes of failure in what it built. The execution model, $I = A(S,P)$, makes the portable artifact a specification and the platform-specific artifact a deployment-time product.

\label{sec:disc:future}
One boundary organizes the system and its honesty: operating-system visibility does not guarantee semantic identifiability. What the platform or the user supplies is meaning; what the agent materializes is infrastructure. Where that meaning comes from is the question we would take up next: a \emph{zero-or-one-query} policy, asking the user a single question wherever the evidence under-determines a device, would close the gap by a change of delivery rather than of mechanism. Compilers once let developers state intent and left processor-specific instruction to the machine. If hardware integration is undergoing the same transition, the artifact worth shipping is no longer the integration but the specification that produces it, and the runtime worth trusting is an agent that can read that specification on whatever machine it lands on, build what the machine needs, and keep it alive. In the deployments this paper reports, that is already the arrangement: the living coding agent was the software; the code was just the artifact.

\bibliographystyle{IEEEtran}
\bibliography{related-work}

@inproceedings{liang2023cap,
  title={Code as policies: Language model programs for embodied control},
  author={Liang, Jacky and Huang, Wenlong and Xia, Fei and Xu, Peng and Hausman, Karol and Ichter, Brian and Florence, Pete and Zeng, Andy},
  booktitle={2023 IEEE International conference on robotics and automation (ICRA)},
  pages={9493--9500},
  year={2023},
  organization={IEEE}
}

@inproceedings{ahn2022saycan,
  author    = {Michael Ahn and others},
  title     = {Do As I Can, Not As I Say: Grounding Language in Robotic Affordances},
  booktitle = {Conference on Robot Learning (CoRL)},
  year      = {2022}
}

@inproceedings{zitkovich2023rt2,
  author    = {Brianna Zitkovich and others},
  title={Rt-2: Vision-language-action models transfer web knowledge to robotic control},
  booktitle = {Conference on Robot Learning (CoRL)},
  pages={2165--2183},
  year={2023},
  organization={PMLR}
}

@inproceedings{kim2024openvla,
  author    = {Moo Jin Kim and others},
  title     = {{OpenVLA}: An Open-Source Vision-Language-Action Model},
  booktitle = {Conference on Robot Learning (CoRL)},
  year      = {2024}
}

@article{li2026roboclaw,
  title={Roboclaw: An agentic framework for scalable long-horizon robotic tasks},
  author={Li, Ruiying and Zhou, Yunlang and Zhu, YuYao and Chen, Kylin and Wang, Jingyuan and Wang, Sukai and Hu, Kongtao and Yu, Minhui and Jiang, Bowen and Su, Zhan and others},
  journal={arXiv preprint arXiv:2603.11558},
  year={2026}
}

@inproceedings{fu2026capx,
  author    = {Letian Fu and Justin Yu and Karim El-Refai and others},
  title     = {{CaP-X}: A Framework for Benchmarking and Improving Coding Agents for Robot Manipulation},
  booktitle = {International Conference on Machine Learning (ICML), PMLR 306},
  year      = {2026}
}

@article{royce2024rosa,
  author  = {Rob Royce and Marcel Kaufmann and Jonathan Becktor and others},
  title   = {Enabling Novel Mission Operations and Interactions with {ROSA}: The Robot Operating System Agent},
  journal = {arXiv preprint arXiv:2410.06472},
  year    = {2025}
}

@article{cardenas2026rosclaw,
  author  = {Irvin Steve Cardenas and Marcus Anthony Arnett and Natalie Catherine Yeo and Lucky Shah and Jong-Hoon Kim},
  title   = {{ROSClaw}: An {OpenClaw} {ROS}~2 Framework for Agentic Robot Control and Interaction},
  journal = {arXiv preprint arXiv:2603.26997},
  year    = {2026}
}

@article{coleman2014moveit,
  author  = {David Coleman and Ioan {\c{S}}ucan and Sachin Chitta and Nikolaus Correll},
  title   = {Reducing the Barrier to Entry of Complex Robotic Software: A {MoveIt!} Case Study},
  journal = {Journal of Software Engineering for Robotics},
  volume  = {5},
  number  = {1},
  year    = {2014}
}

@misc{anthropic2024mcp,
  author       = {{Model Context Protocol contributors}},
  title        = {Model Context Protocol Specification},
  year         = {2025},
  howpublished = {\url{https://modelcontextprotocol.io/specification/2025-11-25}}
}

@inproceedings{yang2025iotmcp,
  author    = {Ningyuan Yang and Guanliang Lyu and Mingchen Ma and others},
  title     = {{IoT-MCP}: Bridging {LLMs} and {IoT} Systems Through Model Context Protocol},
  booktitle = {ACM Workshop on Wireless Network Testbeds, Experimental Evaluation and Characterization (WiNTECH)},
  year      = {2025},
  note      = {arXiv:2510.01260}
}

@article{yang2026dcp,
  author  = {Dongxu Yang},
  title   = {Device Context Protocol: A Compact, Safety-First Architecture for {LLM}-Driven Control of Constrained Devices},
  journal = {arXiv preprint arXiv:2605.26159},
  year    = {2026}
}

@article{lee2025rcp,
  author  = {Lambert Lee and Joshua Lau},
  title   = {Robot Context Protocol ({RCP}): A Runtime-Agnostic Interface for Agent-Aware Robot Control},
  journal = {arXiv preprint arXiv:2506.11650},
  year    = {2025}
}

@article{mastouri2026autompc,
  author  = {Meriem Mastouri and Emna Ksontini and Amine Barrak and Wael Kessentini},
  title   = {From {REST} to {MCP}: An Empirical Study of {API} Wrapping and Automated Server Generation for {LLM} Agents},
  journal = {arXiv preprint arXiv:2507.16044},
  year    = {2026}
}

@article{guan2026roboneuron,
  author  = {Weifan Guan and Qinghao Hu and Huasen Xi and Chenxiao Zhang and Aosheng Li and Jian Cheng},
  title   = {{RoboNeuron}: A Middle-Layer Infrastructure for Agent-Driven Orchestration in Embodied {AI}},
  journal = {arXiv preprint arXiv:2512.10394},
  year    = {2026}
}

@article{macenski2022ros2,
  author  = {Steven Macenski and Tully Foote and Brian Gerkey and Chris Lalancette and William Woodall},
  title   = {Robot Operating System 2: Design, Architecture, and Uses in the Wild},
  journal = {Science Robotics},
  volume  = {7},
  number  = {66},
  year    = {2022}
}

@inproceedings{ryzhyk2014termite,
  author    = {Leonid Ryzhyk and Adam Walker and John Keys and Alexander Legg and Arun Raghunath and Michael Stumm and Mona Vij},
  title     = {User-Guided Device Driver Synthesis},
  booktitle = {USENIX Symposium on Operating Systems Design and Implementation (OSDI)},
  year      = {2014}
}

@inproceedings{yang2024sweagent,
  author    = {John Yang and others},
  title     = {{SWE-agent}: Agent-Computer Interfaces Enable Automated Software Engineering},
  booktitle = {NeurIPS},
  year      = {2024}
}

@inproceedings{jana2026terraformer,
  author    = {Prithwish Jana and others},
  title     = {{TerraFormer}: Automated Infrastructure-as-Code with {LLMs} Fine-Tuned via Policy-Guided Verifier Feedback},
  booktitle = {ICSE-SEIP},
  year      = {2026}
}

@article{stoica2024specs,
  author  = {Ion Stoica and Matei Zaharia and Joseph Gonzalez and Ken Goldberg and Koushik Sen and Hao Zhang and others},
  title   = {Specifications: The Missing Link to Making the Development of {LLM} Systems an Engineering Discipline},
  journal = {arXiv preprint arXiv:2412.05299},
  year    = {2024}
}

@article{kephart2003autonomic,
  author  = {Jeffrey O. Kephart and David M. Chess},
  title   = {The Vision of Autonomic Computing},
  journal = {IEEE Computer},
  volume  = {36},
  number  = {1},
  pages   = {41--50},
  year    = {2003}
}

@inproceedings{bouzenia2025repairagent,
  author    = {Islem Bouzenia and Premkumar Devanbu and Michael Pradel},
  title     = {{RepairAgent}: An Autonomous, {LLM}-Based Agent for Program Repair},
  booktitle = {ICSE},
  year      = {2025}
}

@inproceedings{mei2025aios,
  author    = {Kai Mei and Xi Zhu and Wujiang Xu and others},
  title     = {{AIOS}: {LLM} Agent Operating System},
  booktitle = {Conference on Language Modeling (COLM)},
  year      = {2025}
}

@article{packer2023memgpt,
  author  = {Charles Packer and others},
  title   = {{MemGPT}: Towards {LLMs} as Operating Systems},
  journal = {arXiv preprint arXiv:2310.08560},
  year    = {2023}
}

@inproceedings{guo2025sensormcp,
  author    = {Yunqi Guo and Guanyu Zhu and Kaiwei Liu and Guoliang Xing},
  title     = {{SensorMCP}: A Model Context Protocol Server for Custom Sensor Tool Creation},
  booktitle = {ACM International Conference on Mobile Systems, Applications and Services (MobiSys)},
  pages     = {747--752},
  year      = {2025},
  doi       = {10.1145/3711875.3736687}
}

@inproceedings{liu2025tasksense,
  author    = {Kaiwei Liu and others},
  title     = {{TaskSense}: A Translation-like Approach for Tasking Heterogeneous Sensor Systems with {LLMs}},
  booktitle = {ACM Conference on Embedded Networked Sensor Systems (SenSys)},
  year      = {2025},
  doi       = {10.1145/3715014.3722070}
}

@article{li2026skilled,
  author  = {Yiming Li and Yuhan Cheng and Mingchen Ma and others},
  title   = {Skilled {AI} Agents for Embedded and {IoT} Systems Development},
  journal = {arXiv preprint arXiv:2603.19583},
  year    = {2026}
}

\end{document}